\documentclass{article} 
\usepackage{paper,times}
\usepackage{graphicx}
\usepackage{tabularx}
\usepackage{booktabs}
\usepackage{makecell}
\usepackage{array}
\usepackage{caption}
\usepackage{subcaption}
\usepackage{ulem}
\usepackage{mathtools}

\usepackage{amsmath,amsfonts,bm}

\def\eqref#1{equation~\ref{#1}}

\def\1{\bm{1}}

\DeclareMathAlphabet{\mathsfit}{\encodingdefault}{\sfdefault}{m}{sl}
\SetMathAlphabet{\mathsfit}{bold}{\encodingdefault}{\sfdefault}{bx}{n}

\usepackage{hyperref}
\usepackage{url}

\iclrfinalcopy 

\begin{document}

\begin{center}
{\LARGE \bfseries Cosmos-Surg-dVRK: World Foundation Model-based Automated Online Evaluation of Surgical Robot Policy Learning \\[1.5ex]}

\textbf{Lukas Zbinden}$^{*1}$ \quad \textbf{Nigel Nelson}$^{*1}$ \quad \textbf{Juo-Tung Chen}$^{2}$ \quad \textbf{Xinhao Chen}$^{2}$ \\[0.2em]
\textbf{Ji Woong (Brian) Kim}$^{2,3}$ \quad \textbf{Mahdi Azizian}$^{1}$ \quad \textbf{Axel Krieger}$^{2}$ \quad \textbf{Sean Huver}$^{1}$ \\
$^{1}$NVIDIA \quad $^{2}$Johns Hopkins University \quad $^{3}$Stanford University
\end{center}

\begingroup
  \renewcommand\thefootnote{}
  \footnotetext{\textsuperscript{*}Equal contribution. Correspondence to: lzbinden@nvidia.com, nigeln@nvidia.com}
  \addtocounter{footnote}{0}
\endgroup

\begin{figure}[ht]
    \centering
    \begin{subfigure}[t]{0.47\linewidth}
        \centering
        \includegraphics[width=\linewidth]{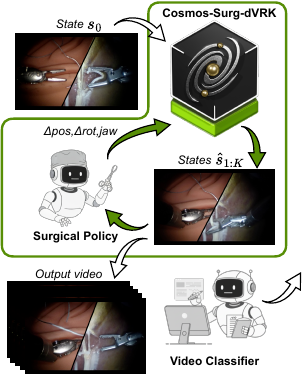}
        \caption{System overview}
        \label{fig:cosmos-policy-eval-overview-a}
    \end{subfigure}
    \hspace{0pt} 
    \begin{subfigure}[t]{0.51\linewidth}
        \centering
        \includegraphics[width=\linewidth]{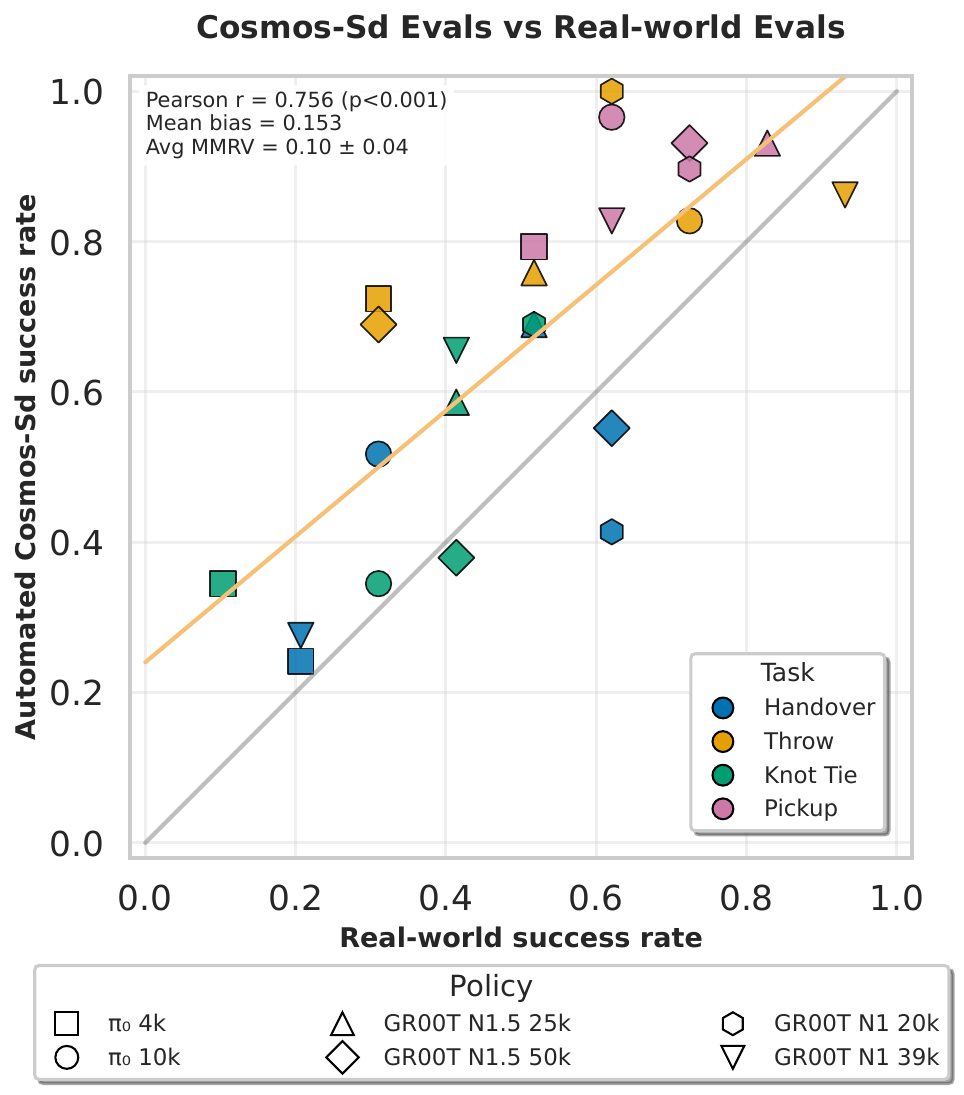}
        \caption{Automated evaluation}   
        \label{fig:cosmos-policy-eval-overview-b}
    \end{subfigure}

    \caption{\textbf{Online evaluation of surgical robot policies in Cosmos-Surg-dVRK simulation.} We propose a framework for automated policy evaluation using Cosmos-Surg-dVRK, a Cosmos world foundation model (WFM) finetune, to perform simulated surgical policy rollouts and subsequent automated success rate evaluation using a video classifier. Evaluation proceeds by initializing the policy with an observed frame at state $s_0$. 
    Conditioned on $s_0$, the policy and Cosmos-Surg-dVRK then generate $K$ action-conditioned future frames autoregressively, $s_{1:K}$.
    In each iteration, the predicted frames are appended to the output video, and the last predicted frame is used as input to the surgical policy for the next iteration. After rollout completion, the generated video is automatically labeled for task success or failure using a trained video classifier, enabling objective selection of the most promising surgical policies for real-robot evaluation and deployment.}
    \label{fig:cosmos-policy-eval-overview}
\end{figure}

\begin{abstract}
The rise of surgical robots and vision-language-action models has accelerated the development of autonomous surgical policies and efficient assessment strategies. However, evaluating these policies directly on physical robotic platforms such as the da Vinci Research Kit (dVRK) remains hindered by high costs, time demands, reproducibility challenges, and variability in execution. World foundation models (WFM) for physical AI offer a transformative approach to simulate complex real-world surgical tasks, such as soft tissue deformation, with high fidelity. This work introduces Cosmos-Surg-dVRK, a surgical finetune of the Cosmos WFM, which, together with a trained video classifier, enables fully automated online evaluation and benchmarking of surgical policies. We evaluate Cosmos-Surg-dVRK using two distinct surgical datasets. On tabletop suture pad tasks, the automated pipeline achieves strong correlation between online rollouts in Cosmos-Surg-dVRK and policy outcomes on the real dVRK Si platform, as well as good agreement between human labelers and the V-JEPA 2-derived video classifier. Additionally, preliminary experiments with ex-vivo porcine cholecystectomy tasks in Cosmos-Surg-dVRK demonstrate promising alignment with real-world evaluations, highlighting the platform's potential for more complex surgical procedures. 
\end{abstract}

\section{Introduction}
\label{sec:intro}
World models have emerged as a foundational approach for enabling intelligent agents to understand and act in complex simulated environments. Building on early work by \citet{ha2018worldmodels}, they learn compact latent representations of spatial and temporal dynamics and predict the consequences of actions in context. Leveraging diffusion processes, recent efforts have introduced large-scale multi-modal world foundation models (WFMs)~\citep{Wan2025OpenLargeScaleVideoGen,Agarwal2025CosmosWFM,kong2024hunyuanvideo,xie2025sana,genie3}. These video generative models serve as scalable, general-purpose learned simulators that encode and synthesize diverse physical phenomena, approximate scene dynamics, render plausible sensory observations, and facilitate policy evaluation and training, enabling progress on sim-to-real transfer. 

At the intersection of these advances lies the burgeoning field of physical AI, AI systems equipped with sensors and actuators that enable perception, reasoning, and actuation in the physical world. Robotics, as a core subfield of physical AI, has benefited from the integration of vision-language models (VLMs), generative policies, and world model-based reasoning. State-of-the-art approaches increasingly rely on learned simulators and predictive models not only to improve sample efficiency, but also to foster generalization and robustness under distribution shift and real-world variability.

The Cosmos World Foundation Model (Cosmos WFM) ~\citep{Agarwal2025CosmosWFM} is a video-based world model intended to support scalable simulation and decision-making for physical AI. It combines large curated video datasets with transformer-based diffusion and autoregressive components, together with video tokenization. In practice, Cosmos WFM can be used to construct learned simulators, either generalist or domain-specific digital twins of the physical world. 

These advancements are increasingly relevant to autonomous surgical robotics, a subdomain of physical AI. The drive for reliable automation in surgical settings is motivated by precision, safety, reproducibility, and an anticipated shortage of surgeons~\citep{AAMC2024PhysicianSupplyDemand}, yet is constrained by the cost, time, and risk of evaluating policies on real hardware. The da Vinci Research Kit (dVRK) illustrates these constraints: data collection and iterative policy refinement on the robot are resource-intensive and may raise safety concerns.

In this work, we investigate the application of Cosmos WFM as a learned simulator for surgical robot policy evaluation, introducing Cosmos-Surg-dVRK. By shifting surgical policy evaluation from direct dVRK Si trials to a Cosmos-Surg-dVRK-simulated environment, we aim to demonstrate a paradigm for efficient, safe, and reproducible benchmarking of surgical robot policies entirely in simulation. This approach has the potential to accelerate development cycles and reduce operational costs by enabling novel learning algorithms to be tested and iterated entirely in a domain-specific digital twin that replicates real procedures.

\subsection*{Motivation and Problem Statement}
One major challenge in advancing vision-language-action (VLA) models for surgical autonomy, also known as surgical robot policies, is the substantial time and effort required to evaluate newly trained policies. Traditionally, two approaches exist for this evaluation. First, researchers with physical access to robotic hardware and surgical assets, such as cadaver or animal labs, deploy trained models to assess policy success and failure. Second, the researchers emulate the real-world surgical task to a sufficient degree in simulation, allowing them to approximate real-world performance by evaluating policy outcomes virtually. Surgical robotics poses a unique challenge for both approaches.

\textbf{Real-world evaluations} provide gold standard metrics but have several limitations:

\begin{enumerate}
  \item Regulatory Hurdles: Evaluating policies ex-vivo or in-vivo requires navigating surgical asset management, ethics committees, institutional review boards, liability, insurance, privacy laws, and more.
  \item Reproducibility: Variability in cable-actuated surgical robots, differences between robot generations, and challenges in replicating arm and endoscope positioning hinder consistent replication of surgical policy performance.
  \item Time: Each evaluation demands task-specific calibration of robots and cameras, manual resetting of scene and robot after each rollout, and serial execution due to limited platform availability, preventing parallel testing.
\end{enumerate}

\textbf{Simulation-based evaluations} can address several real-world challenges, but also have drawbacks:
\begin{enumerate}
  \item Soft-Body Dynamics: Even advanced simulators with cutting, suturing, and cauterizing capabilities lack realistic modeling of complex, multi-material tissue properties.
  \item Cost: Accurately emulating real-world physics and appearance demands significant engineering and setup time.
\end{enumerate}
Ultimately, real-world evaluation remains essential as a final validation step. However, no generalizable, low-cost solution currently exists for faithful surgical policy evaluation during development, due to the challenges posed by soft-body dynamics in state-of-the-art simulators.

\begin{figure}[ht]                
  \centering
  \includegraphics[width=0.99\linewidth]{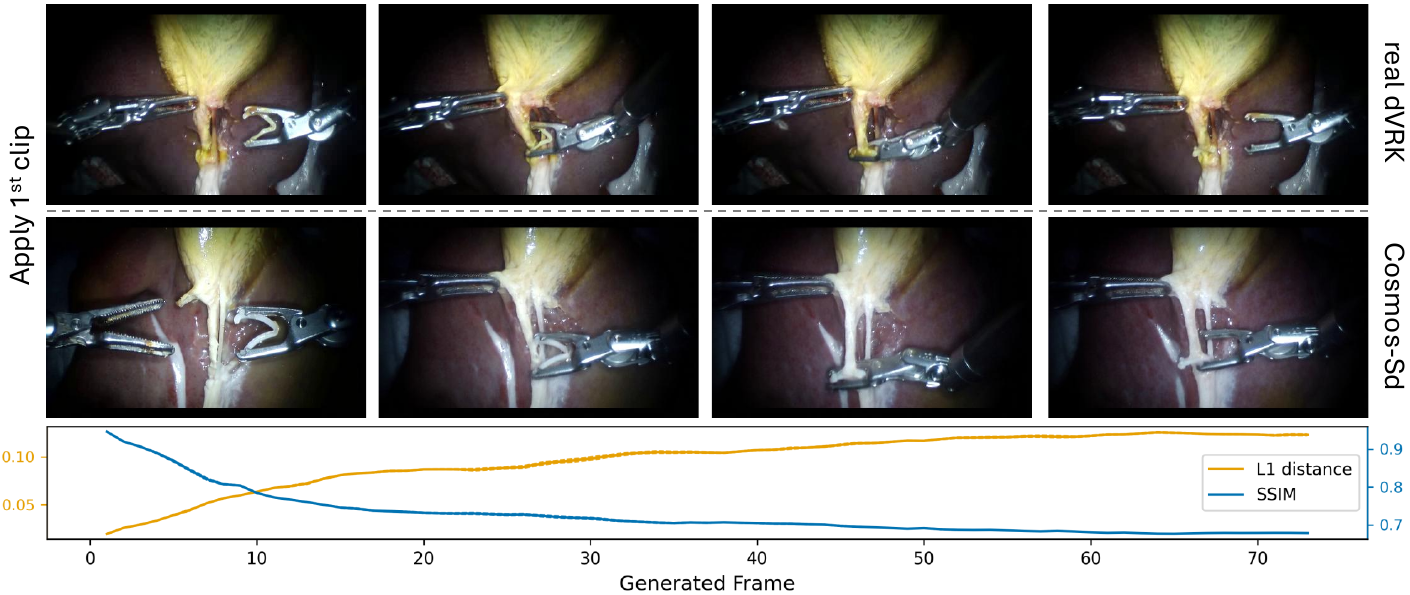}
  \caption{
    \textbf{Soft tissue simulation with Cosmos-Surg-dVRK.} Surgical  simulation in ex-vivo porcine cholecystectomy. Top row: example as recorded on the real dVRK Si. Middle row: Cosmos-Surg-dVRK generated example with identical kinematic action trajectory. Bottom row: Holdout L1 and SSIM vs. number of generated frames.
  }
  \label{fig:overview_chole_soft_tissue_sim}
\end{figure}

\subsection*{Proposed Approach and Contributions}
We propose Cosmos-Surg-dVRK\footnote{denoted as Cosmos-Sd in figures}, a surgical finetune of the Cosmos WFM, as a software-in-the-loop solution for efficient and cost-effective evaluation of dVRK Si-based surgical robot policies in simulation. Finetuning the Cosmos-Predict2 variant that conditions on kinematic actions achieves two key benefits: (1) Training Cosmos WFM to predict future frames from the current state and predicted actions enables implicit modeling of the robot's kinematic chain, removing the need for explicit articulation or kinematics specification; (2) Post-training an existing Cosmos WFM leverages broad pretraining for physics understanding, while additional finetuning on surgical data allows the model to learn domain-specific tissue and robot interactions. Task-specific soft tissue simulation is thus derived directly from surgical datasets, eliminating manual tuning of physics parameters typical in traditional simulators (Figure~\ref{fig:overview_chole_soft_tissue_sim}).

To our knowledge, no surgery-specific WFM finetunes exist that are directly conditioned on kinematic actions as described. Most current action-conditioned WFM research focuses on domains like video game emulation, automotive, or general robotics. Accordingly, the main contributions of this work are:

\begin{enumerate}
\item We demonstrate that Cosmos-Surg-dVRK facilitates online evaluation of surgical policies, yielding results that strongly correlate with real-world policy performance on tabletop tasks.
\item We show good agreement between the video classifier and human labelers, establishing a fully automated policy evaluation pipeline with strong correlation to real-world outcomes.
\item We extend Cosmos-Surg-dVRK to online policy evaluation of ex-vivo porcine cholecystectomy, demonstrating its ability to model complex real-world tissue deformations.
\item We present a preliminary ablation study underscoring the role of failure episodes in the learning process for surgical task simulation.
\end{enumerate}

\section{Related work}
\paragraph{Vision-language-action (VLA) Foundation Models.}
Recent years have seen the emergence of VLA foundation models as a general framework for robotic intelligence~\citep{sapkota2025visionlanguageactionmodelsconceptsprogress}. These models integrate visual perception, natural language comprehension, and action generation, enabling robots to understand instructions, perceive their environment, and act accordingly with robustness and adaptability. We subsequently introduce the VLAs used in this work. 

$\pi_0$~\citep{pi0_vla} is a scalable, open-world VLA foundation model for generalist robotic manipulation, trained on web, synthetic, and real robot data across language, image, and action modalities. The model demonstrates robust zero-shot generalization to diverse real-robot manipulation tasks. GR00T N1~\citep{nvgr00tn1} is a dual-system VLA foundation model for humanoid robots, featuring a vision-language model backbone for high-level reasoning and a diffusion transformer for motor actions. Trained on web-scale video, synthetic trajectories, and real robot demonstrations, N1 shows competitive performance on multi-modal language-conditioned tasks. Its extension, GR00T N1.5~\citep{nvgr00tn1_5}, advances cross-embodiment learning with scaled parameters and training data, demonstrating improved manipulation and navigation performance across robot bodies. 

\paragraph{Surgical Robot Policy Learning.}
Recent advances in autonomous surgical robotics center on data-driven policy learning, notably imitation learning and transformer architectures~\citep{Kim2024SRT}. Robust imitation policies for the dVRK Si robot have been trained using hybrid-relative motion representations, which address hardware inaccuracies. Building on this, SRT-H~\citep{Kim2025SRTH} introduces a hierarchical transformer framework for language-conditioned planning and visuomotor control, validated on ex-vivo cholecystectomy procedures. These works demonstrate comprehensive autonomy in complex surgical workflows, achieving self-correction, scalability across anatomical variations, and generalization to out-of-distribution scenarios. The SRT and SRT-H series set new benchmarks in scale (16,000+ trajectories) and reliability for contact-rich, long-horizon robotic manipulation.

\paragraph{Robot Policy Evaluation.}
Evaluating robot policies across diverse tasks and hardware remains a major challenge as model complexity increases. Simulation-based approaches, such as SIMPLER~\citep{li24simpler}, allow real-world policies to be assessed in virtual environments, but configuring these simulators for new tasks or embodiments can be resource-intensive. Recent frameworks like WorldEval~\citep{Li2025WorldEval} address these limitations by leveraging learned world models to generate accurate, policy-conditioned video rollouts for scalable, policy-agnostic evaluation. This enables efficient comparison, policy ranking, and automatic detection of failures using multi-modal or statistical criteria.

\subsection*{Limitations of Prior Work}
Prior works in surgical policy evaluation are built on physics-based simulation approaches that employ mathematical modeling to enforce known physical laws. SurgicAI~\citep{wu2024surgicaifinegrainedplatformdata} introduces a platform for surgical robot policy learning supporting dVRK with deformable thread simulation based on continuum mechanics. SurRoL~\citep{Yang2024SurRol} incorporates a material point method simulator for soft body dynamics, while ORBIT-Surgical~\citep{orbit_surgical} provides finite element method-based deformable body simulation. These physics-constrained approaches require explicit material property specification, limiting their adaptability to diverse surgical scenarios and tissue variability. Our work, Cosmos-Surg-dVRK, addresses this gap by learning soft tissue dynamics directly from surgical data distributions, enabling coverage of complex deformable simulation spaces.

In the general policy domain, data-driven world model approaches have been used to learn dynamics patterns from observational data distributions rather than enforcing explicit physical constraints. WorldEval~\citep{Li2025WorldEval} employs latent action embeddings for world model video generation, requiring access to the policy's internal representations. In contrast, we employ a standard relative Cartesian action space, enabling a plug-and-play interface that mirrors the API of a real robot controller. Other approaches, leveraging physics-based simulators~\citep{li24simpler} and WFMs~\citep{worldgym}, also adopt this intuitive action space approach, but focus on more general rigid-body manipulation tasks.


\section{Cosmos-Surg-dVRK: Cosmos WFM for Automated Surgical Robot Policy Evaluation}
In this section, we first outline the surgical policy evaluation task that builds on Cosmos-Surg-dVRK, our kinematic action-conditioned surgical simulation platform based on Cosmos WFM. We then propose the V-JEPA 2-based classifier for automated policy rollout evaluation. A system overview is presented in Figure~\ref{fig:cosmos-policy-eval-overview-a}.

\subsection{Overview of Approach}
Given a collection of multi-task surgical policies, represented either by distinct models or by different training regimes, this study aims to benchmark their success rate via online rollouts in simulation, thereby eliminating the need for immediate evaluation on a physical robot platform. The objective is to identify the policy exhibiting the highest simulated performance prior to deployment on the physical robot. To assess the validity of the Cosmos-Surg-dVRK simulation framework, a set of policies are trained first and then evaluated on the dVRK Si platform, recording both the robot's initial state (captured as the first video frame) and the observed success or failure for each policy rollout. Subsequently, these same policies are rolled out online in the Cosmos-Surg-dVRK environment, using the recorded initial states from the dVRK Si as standardized starting conditions. The success rates attained in both the simulation and physical settings are then compared through statistical analysis to determine the fidelity of the simulation framework in replicating real-world performance.

\subsection{V-JEPA 2 and Attentive Probing: Automated policy rollout evaluation}
While Cosmos-Surg-dVRK generates complete videos for each policy rollout, facilitating manual evaluation, reviewing hundreds of videos remains a time-consuming process, particularly when distinguishing subtle decision boundaries between success and failure. As such, we utilize V-JEPA 2~\citep{assran2025vjepa2selfsupervisedvideo} for its strong pretrained video features and train an attentive probe classifier to automatically label success and failure events on our Cosmos-Surg-dVRK rollouts. We train a single classifier for all tabletop tasks using a frozen V-JEPA 2 ViT-H backbone on a manually collected dataset of 2,310 video clips. To accommodate V-JEPA 2's limited context window, we process each rollout in overlapping video chunks. Then, the complete rollout is labeled as a success or failure based on whichever of those outcomes appears first. Additional implementation details for the V-JEPA 2 classifier are provided in \ref{app:vjepa_implementation_details}.

\section{Implementation}
\subsection{Implementation details}
For both the tabletop and cholecystectomy datasets, we separately finetune the action-conditional Cosmos WFM model Cosmos-Predict2-2B-Video2World~\citep{nvidia_cosmos_predict2} with video and kinematic action paired data sampled at 10 Hz for 20,000 steps on 32x A100 GPUs. For training and inference, at each iteration \(i\), Cosmos WFM takes the current
state \(s_i\) and the policy-originated action sequence \(a_{i:i+K-1}=(a_i,\dots,a_{i+K-1})\) with \(K=12\),
and predicts the subsequent \(K\) frames \(\hat{s}_{i+1:i+K}\). For the next iteration \(i' = i+K\), we set \(s_{i'} \coloneqq \hat{s}_{i+K}\), so the model rolls forward autoregressively. For both datasets, the Cosmos WFM model is trained on all available tasks. The training configuration is listed in Table~\ref{tab:training_config}.

\subsubsection{Surgical policy rollout in Cosmos-Surg-dVRK}
For the surgical policy rollout in Cosmos-Surg-dVRK, we use an online evaluation loop. Each policy runs on one NVIDIA A100 GPU, communicating via socket with Cosmos-Surg-dVRK, running on two NVIDIA A100 GPUs. We limit the number of evaluation steps to 1000 for needle throw, and 750 for needle pickup, needle handover, and knot tying in the tabletop tasks. For cholecystectomy tasks, we use a limit of 500 steps. For each policy and Cosmos-Surg-dVRK checkpoint, we generate rollout videos for each task, with ten trials per task, and run each trial 3 times, each with a distinct random seed for Cosmos-Surg-dVRK generation, to measure variability and robustness. The different policies are executed without random seeding. The policies predict multi-step action chunks of various lengths. Since Cosmos-Surg-dVRK has a prediction horizon of $K = 12$ actions, only the first $K$ policy actions are processed in each iteration. Due to $\pi_0$ having a much larger action chunk, $\pi_0$ is trained at 30Hz while the GR00T-N$x$ policies are trained at 15Hz. At inference time, the predictions of the policies are downsampled to 10Hz to match Cosmos-Surg-dVRK's finetuning, and then only the first $K$ actions are used to condition Cosmos-Surg-dVRK's next-frame generation.

\subsubsection{Surgical policy rollout on dVRK}
We evaluate all policies on the dVRK Si platform under a fixed robot configuration with consistent suture pad placement and varying initial needle position. Policies are executed in real-time with outputs corresponding to relative translation, rotation, and absolute jaw angle. The first 20 actions are executed on the robot before we run the policy inference again. This execution horizon was empirically chosen to balance responsiveness and stability.
All models are evaluated on a dual NVIDIA RTX 4090 workstation. For the tabletop suturing benchmark, we evaluate needle pickup, needle throw, needle handover, and knot tying, with a maximum time limit of one minute per task. Each policy is tested for ten rollouts per task, with the environment manually reset between trials.

For ex-vivo porcine cholecystectomy, we compare against the SRT-H policy results reported in~\citet{Kim2025SRTH}. Since our evaluation setup does not use wrist camera inputs, we specifically reference the “no wrist camera” ablation results from their work, which condition only on the endoscope view, ensuring a fair and consistent comparison.

\subsection{Statistical Analysis} We use various evaluation metrics to quantitatively analyze the relationship  between the policy performance in Cosmos-Surg-dVRK simulation and the real dVRK.

\textbf{Success rate.} To benchmark surgical policies within the simulation framework we use the success rate, defined as the proportion of successful task completions: \(\mathrm{SR} = \frac{S}{N} = \frac{1}{N}\sum_{i=1}^{N} \mathbf{1}\{\text{success}_i\}\), where \(S\) is the number of successes among \(N\) trials.

\medskip
\textbf{Pearson correlation coefficient (Pearson $r$).} 
The Pearson correlation coefficient quantifies the degree and the direction of the linear relationship between two variables and ranges from $-1$ to $1$. A Pearson correlation coefficient close to 1 signifies a highly effective evaluation proxy, indicating that improvements in Cosmos-Surg-dVRK success rates directly correspond to linear increases in real-world success rates. We follow the interpretation by~\citet{Evans1996}.

\medskip
\textbf{Mean maximum rank violation (MMRV).} 
The mean maximum rank violation~\citep{li24simpler} measures the consistency of policy rankings between simulated and real evaluations and ranges from 0 to 1 (lower is better). If the simulator incorrectly ranks policies, it is penalized in MMRV by the corresponding margin in real-world performance.

\medskip
\textbf{Mean bias error (MBE).} 
To quantify bias in success rate evaluation we use mean bias error, defined as $\mathrm{MBE} = \frac{1}{N} \sum_{i=1}^{N} (S_i - R_i)$, where $S_i$ is the simulated success rate, $R_i$ the real success rate, and $N$ is the number of paired evaluations.

\medskip
\textbf{Intraclass correlation coefficient (ICC).} 
To quantify inter-rater agreement for our human and automated labeling, we use the intraclass correlation coefficient ICC(2,1), absolute agreement~\citep{Koo2016ICCGuideline}.


\section{Experiments}
We conduct a series of experiments using Cosmos-Surg-dVRK to assess its validity as a surrogate for the real dVRK Si robotic platform. Our central hypothesis is that policy evaluations performed in simulation can reliably serve as linear predictors of corresponding real-world policy performance.

\subsection{Datasets}
\begin{figure}[ht]
    \centering
    \begin{subfigure}[t]{0.44\linewidth}
        \centering
        \includegraphics[width=\linewidth]{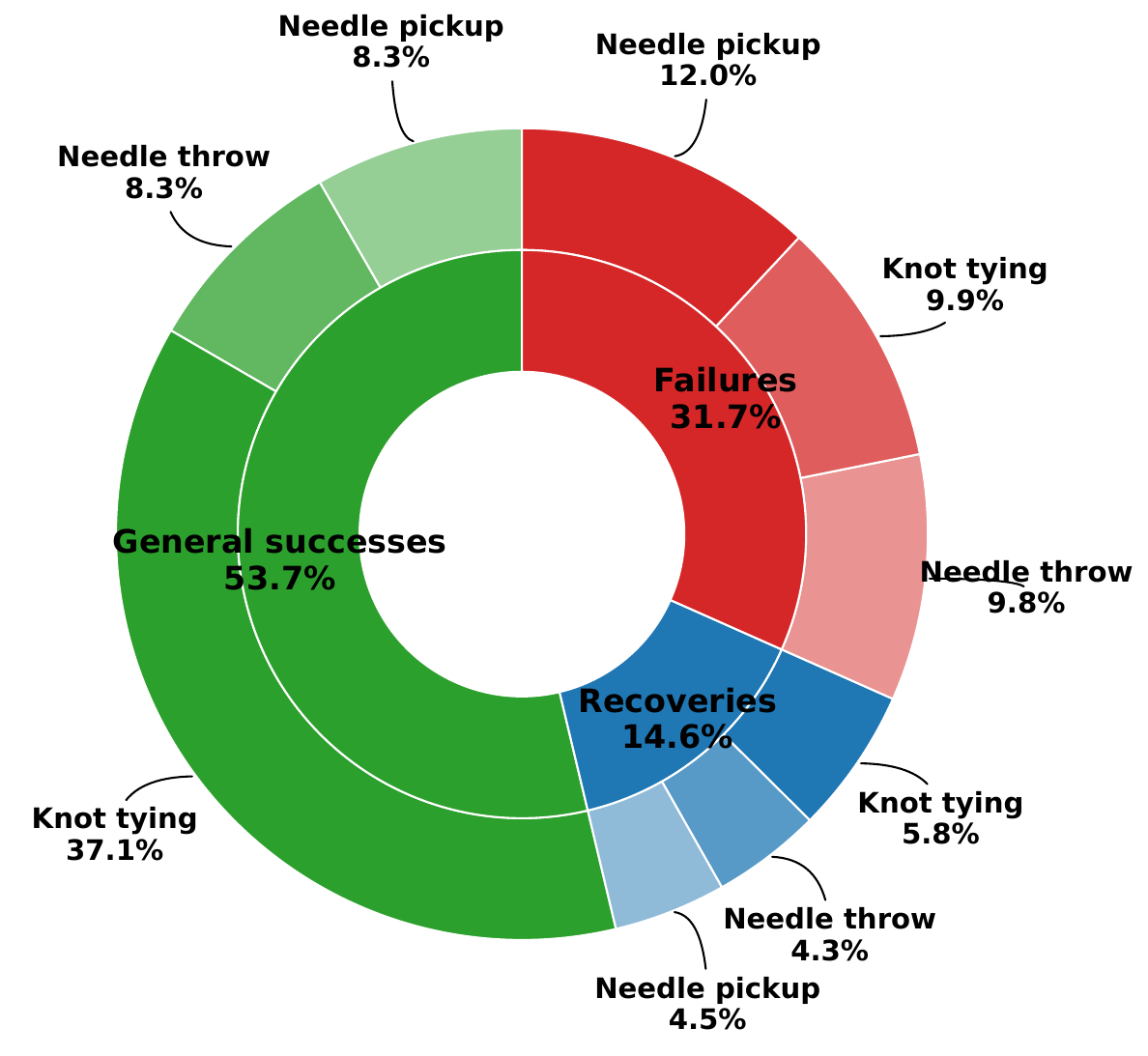}
        \caption{Tabletop surgical tasks}
        \label{fig:tt_ds_composition}
    \end{subfigure}
    \hfill
    \begin{subfigure}[t]{0.48\linewidth}
        \centering
        \includegraphics[width=\linewidth]{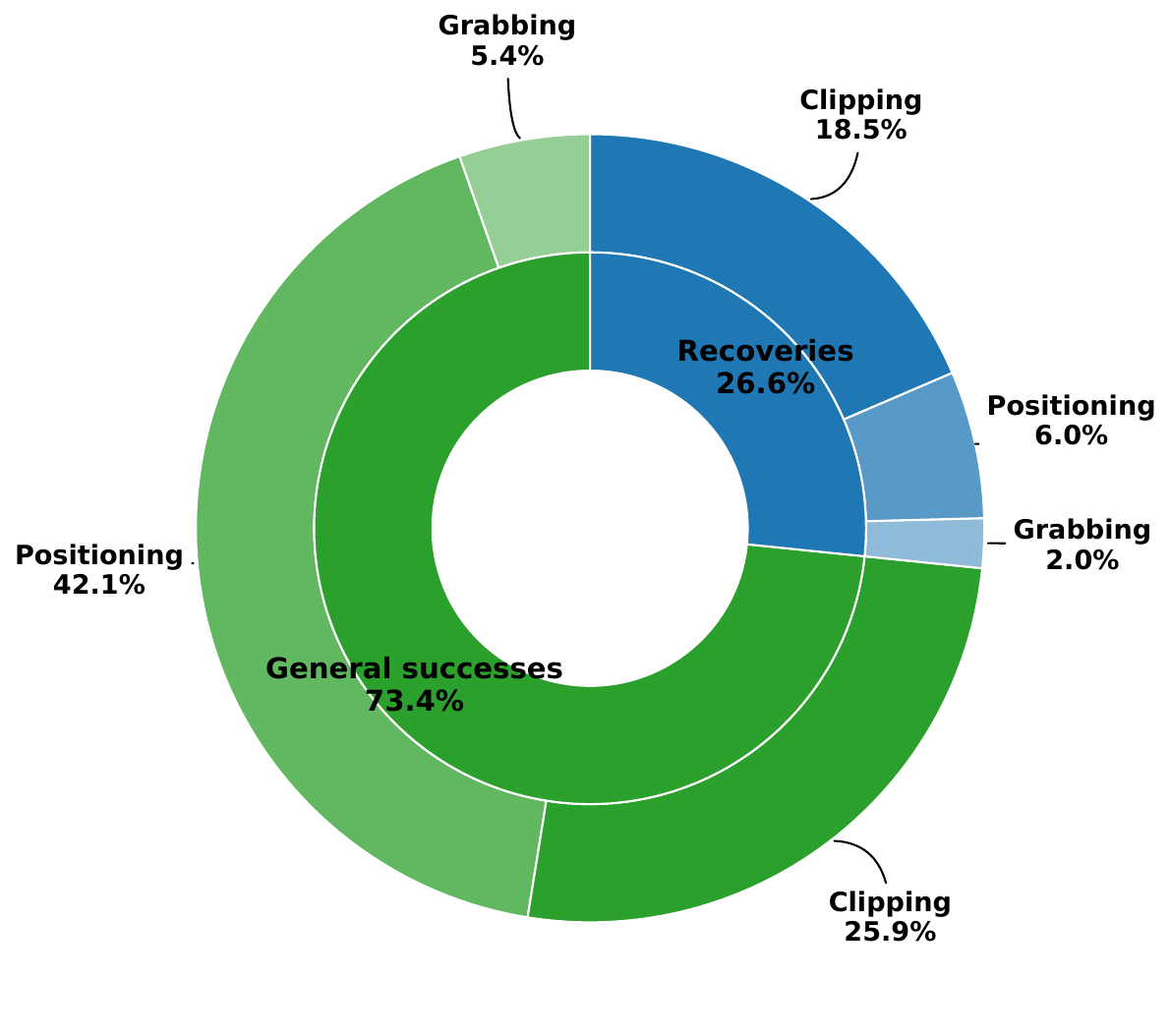}
        \caption{Ex-vivo porcine cholecystectomy}   
        \label{fig:porcine_ds_composition}
    \end{subfigure}

    \caption{\textbf{Surgical autonomy dataset distributions.} 
    a) Tabletop dataset consists of success episodes, recoveries, as well as failure data. The label needle pickup includes task needle handover.
    b) Ex-vivo porcine cholecystectomy dataset consists of success and recovery episodes.}
    \label{fig:jhu_datasets}
\end{figure}

All datasets were recorded at 30Hz on a single dVRK Si system across multiple sessions and expert operators.
The dVRK Si movements were recorded as multi-view video, including endoscope and wrist-camera views, as well as the time-synchronized kinematic sequences. The tabletop suture pad dataset~\citep{haworth2025suturebot} contains four surgical tasks and encompasses 3,036 episodes (Figure~\ref{fig:tt_ds_composition}) with a duration of $\sim$13 hours. The porcine cholecystectomy dataset~\citep{Kim2025SRTH} contains 17 sub-tasks and includes 16,506 episodes (Figure~\ref{fig:porcine_ds_composition}) with a duration of $\sim$18 hours. For pre-processing, we drop redundant consecutive video frames and respective kinematics. 

\subsection{Tabletop Surgical Policy Evaluation in Cosmos-Surg-dVRK versus real dVRK}
We train three different monocular VLAs (policies), $\pi_0$, GR00T N1, and GR00T N1.5, on the dVRK tabletop dataset. For each policy, we evaluate two training regimes, one half training and one full training, using a) Cosmos-Surg-dVRK simulation and b) the real dVRK Si. For each policy checkpoint and environment, the success rate for each surgical task is calculated. The surgical tasks are needle pickup, needle handover, needle throw, and knot tying. Each policy is rolled out in both environments ten times per task (ten trials), each time with a different initial robot state. For each dVRK rollout, a multi-view video with paired kinematics is recorded. The corresponding Cosmos-Surg-dVRK rollout is initialized with the first endoscopic video frame from the dVRK Si recording. In the following we present the results for both manual evaluation and automated evaluation of the simulated policy rollouts.

\subsubsection{Manual Evaluation of Cosmos-Surg-dVRK generated Tabletop policy rollouts}
\label{sec:manual_eval}
The success rates are determined manually both for the dVRK and the Cosmos-Surg-dVRK generated videos. For the generated videos, two raters each label three Cosmos-Surg-dVRK seeds and their averaged scores are collected as the final success rate. 

\begin{figure}[h]                
  \centering
    \includegraphics[width=0.55\linewidth]{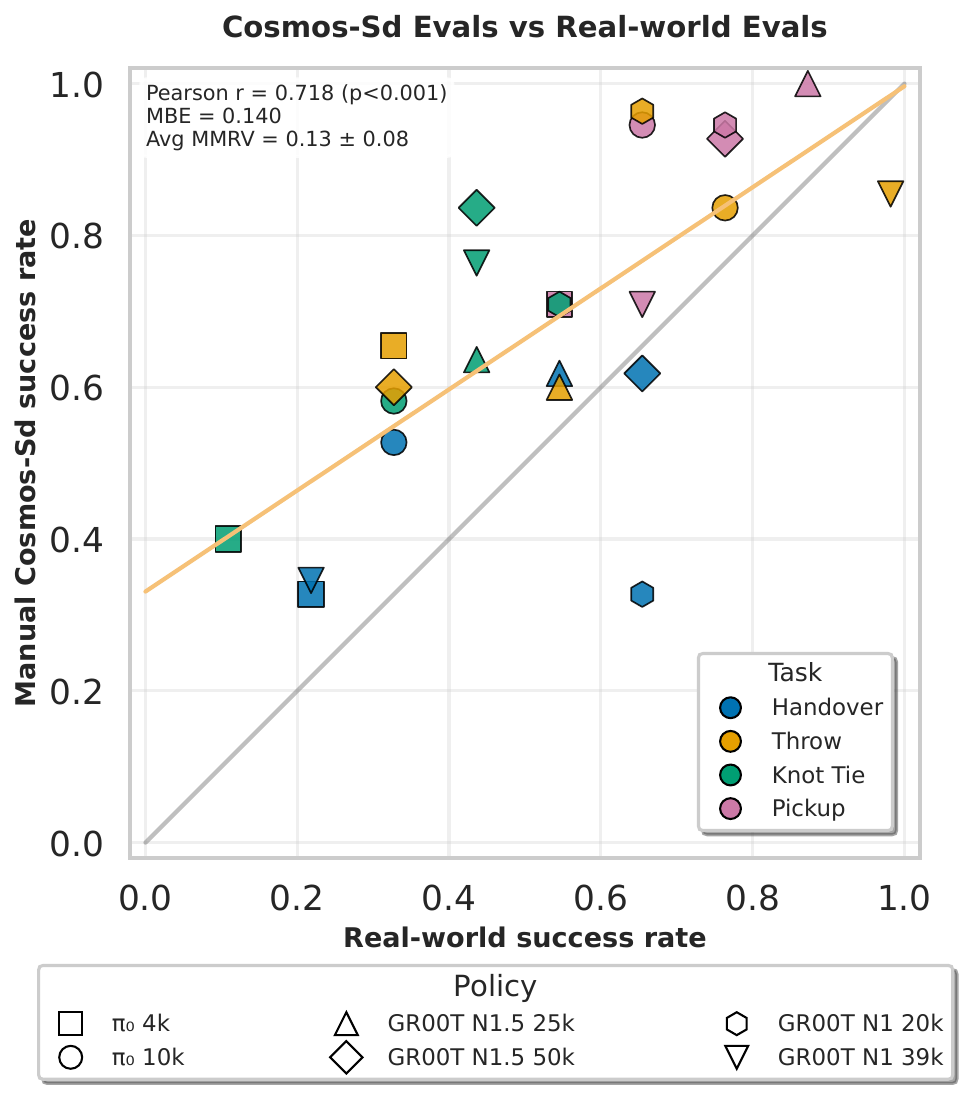}
  \caption{
   \textbf{Manual Cosmos-Surg-dVRK vs. real-world tabletop success rates}. Relationship between manual surgical policy success rate evaluation in Cosmos-Surg-dVRK simulation (vertical axis) and on the real-world dVRK (horizontal axis). Each policy is shown with two training regimes: half–training and full–training across four tabletop suture pad tasks. 
  }
  \label{fig:policy-success}
\end{figure}

\begin{figure}[ht]                
  \centering
  \includegraphics[width=0.99\linewidth]{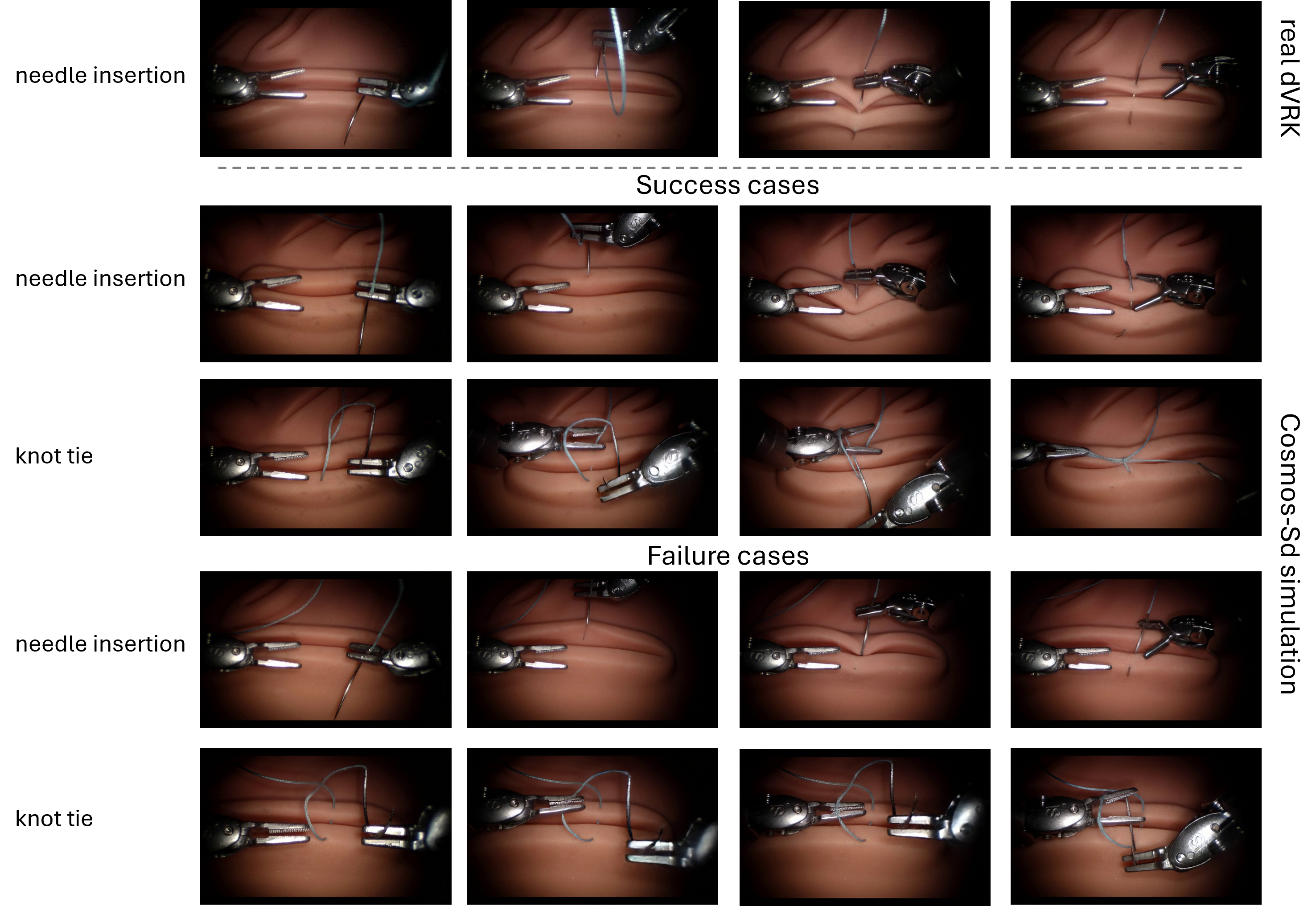}
  \caption{
    \textbf{Qualitative examples of tabletop policy online rollouts in Cosmos-Surg-dVRK.} Each row represents one different policy rollout. Top row: Successful example as recorded on the real dVRK. Middle two rows: Successfully completed tasks in Cosmos-Surg-dVRK. Bottom two rows: Examples of failed tasks in Cosmos-Surg-dVRK. Four frames per task are chosen individually to best represent the example.
  }
  \label{fig:tt-policy-eval-qual-results}
\end{figure}

The two raters achieve an ICC(2,1) of 0.811 in labeling success versus failure across all four tabletop tasks, indicating good reliability. The per-task Pearson and MMRV scores are provided in Table~\ref{tab:policy_eval_quantity_all}, with Cosmos-Surg-dVRK achieving averages of 0.707 (Pearson) and 0.129 (MMRV). Across all tasks and training regimes, policy evaluations in Cosmos-Surg-dVRK display a strong positive correlation with those on the dVRK (Pearson $r=0.718$, $p < 0.001$), as illustrated in Figure~\ref{fig:policy-success}. Despite this alignment, Cosmos-Surg-dVRK rollouts exhibit a positive bias, with policies tending to achieve higher average success rates than on the real dVRK (see Table~\ref{tab:mbe}), with an MBE of 0.140. Qualitative results of online rollouts in Cosmos-Surg-dVRK are shown in Figure~\ref{fig:tt-policy-eval-qual-results}. 

\begin{table*}[t]
    \caption{\textbf{Quantitative results for policy evaluation in Cosmos-Surg-dVRK simulation.}. Pearson correlation coefficient (higher is better) and MMRV (Mean Maximum Rank Violation; lower is better) between Cosmos-Surg-dVRK and dVRK Si policy evaluations. Manual refers to human-labeled task outcomes in the Cosmos-Surg-dVRK simulation, while automated refers to evaluation scores predicted by the V-JEPA 2 classifier.
    }
    \vspace{-4pt}
    \centering
    \small 
    \renewcommand{\arraystretch}{0.6}
    \setlength{\tabcolsep}{0pt} 
    \begin{tabularx}{\textwidth}{l *{10}{>{\centering\arraybackslash}X}}
        \toprule
        \textbf{Evaluation method} & \multicolumn{2}{c}{\textbf{Handover}} & \multicolumn{2}{c}{\textbf{Throw}} & \multicolumn{2}{c}{\textbf{Knot Tie}} & \multicolumn{2}{c}{\textbf{Pickup}} & \multicolumn{2}{c}{\textbf{Avg.}} \\
        \cmidrule(lr){2-3} \cmidrule(lr){4-5} \cmidrule(lr){6-7} \cmidrule(lr){8-9} \cmidrule(lr){10-11}
        & {Pearson} $\uparrow$ & {MMRV} $\downarrow$
        & {Pearson} $\uparrow$ & {MMRV} $\downarrow$
        & {Pearson} $\uparrow$ & {MMRV} $\downarrow$
        & {Pearson} $\uparrow$ & {MMRV} $\downarrow$
        & {Pearson} $\uparrow$ & {MMRV} $\downarrow$ \\
        \midrule
        \makecell[l]{\texttt{Manual}} 
    & 0.468 & 0.217 & 0.716 & 0.183 & 0.840 & 0.050 & 0.806 & 0.067 & 0.707 & 0.129 \\
        \makecell[l]{{\quad w/o failure episodes}} 
            & 0.313 & 0.183 & 0.533 & 0.317 & 0.922 & 0.017 & 0.701 & 0.067 & 0.617 & 0.146 \\
        \midrule
        \makecell[l]{\texttt{Automated}} 
            & 0.656 & 0.133 & 0.639 & 0.117 & 0.729 & 0.033 & 0.639 & 0.100 & 0.666 & 0.096 \\
        \bottomrule
    \end{tabularx}
    \label{tab:policy_eval_quantity_all}
\end{table*}

\begin{table}[h]
\centering
\caption{\textbf{Mean bias error in success rates across Cosmos-Surg-dVRK policy evaluation methods.} MBE, average Pearson, and average MMRV across all tasks are shown. The positive bias in success rates is markedly higher for the Cosmos-Surg-dVRK finetune without failure episodes.}
\label{tab:mbe}
\small
\begin{tabular}{lccc}
\toprule
\textbf{Evaluation Method} & \textbf{MBE (95\% CI)} $\downarrow$ &  \textbf{Pearson} $\uparrow$ & \textbf{MMRV} $\downarrow$ \\
\midrule
        \makecell[l]{\texttt{Manual}} 
     & 0.140 [0.081, 0.199] & 0.71 $\pm$ 0.17 & 0.13 $\pm$ 0.08 \\
        \makecell[l]{{\quad w/o failure episodes}} 
            & 0.325 [0.262, 0.388] & 0.62 $\pm$ 0.26 & 0.15 $\pm$ 0.13 \\
        \midrule
        \makecell[l]{\texttt{Automated}} 
            & 0.153 [0.093, 0.213] & 0.67 $\pm$ 0.04 & 0.10 $\pm$ 0.04 \\
\bottomrule
\end{tabular}
\end{table}


\subsubsection{Automated Evaluation of Cosmos-Surg-dVRK generated Tabletop policy rollouts}
To validate the proposed V-JEPA 2 classifier, we first compare its results with the human labelers as the gold standard and analyze the level of agreement and correlation. Then we proceed to a direct comparison with the real-world success rates.

We generate the V-JEPA 2 classifier labels by running inference on the same set of Cosmos-Surg-dVRK policy rollout videos used in Section~\ref{sec:manual_eval}. For our final V-JEPA 2 evaluations, we average results for each policy-task combination across three generated seeds, following the same procedure as with the human labeling. 

\begin{figure}[h]                
  \centering
  \includegraphics[width=0.55\linewidth]{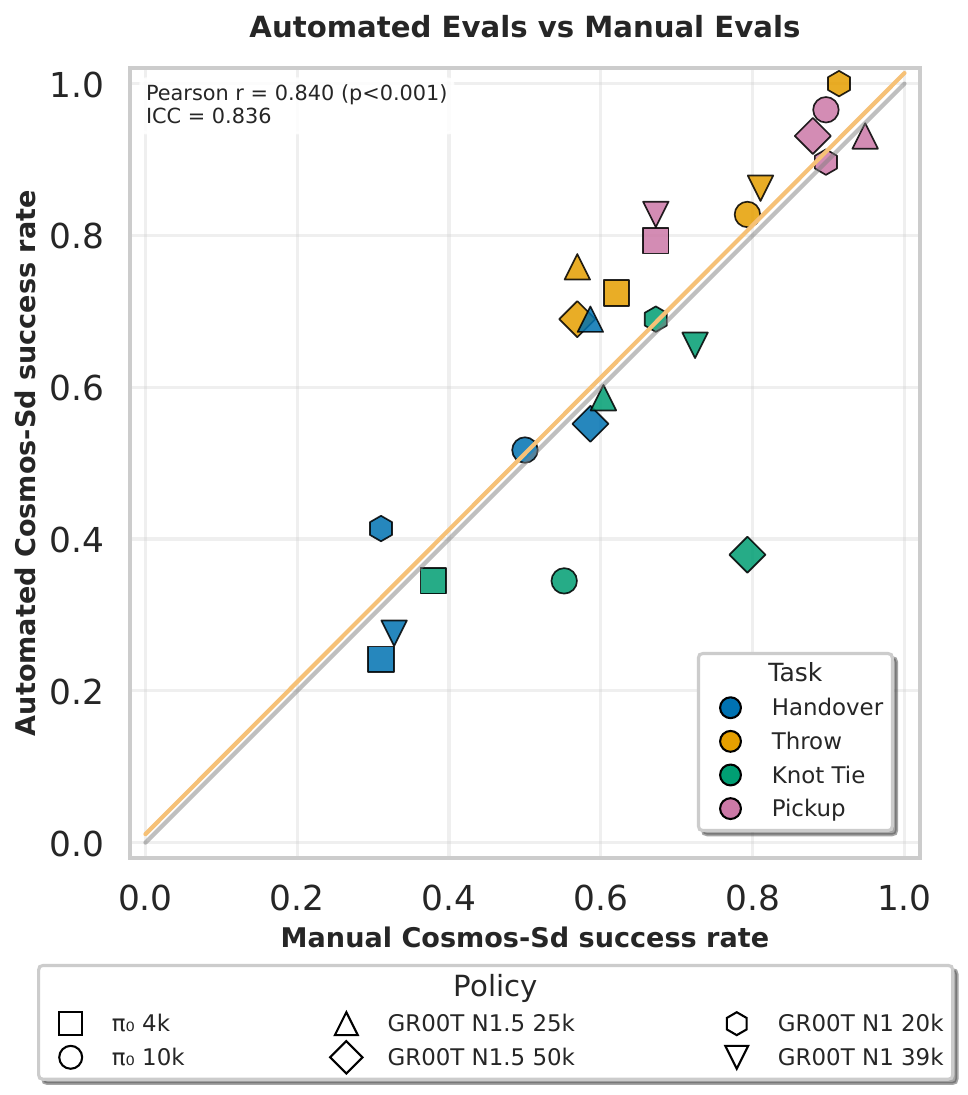}
  \caption{
    \textbf{Comparison of automated vs. manual evaluations.} 
    Relationship between surgical policy success rate evaluations from the V-JEPA 2 classifier and human raters in Cosmos-Surg-dVRK simulation.
  }
  \label{fig:tt-policy-eval-qual-results-vjepa}
\end{figure}

As shown in Figure~\ref{fig:tt-policy-eval-qual-results-vjepa}, comparison of the classifier and human labels on the same test videos yields an ICC(2,1) of 0.836. This reflects good inter-rater agreement, matching or exceeding that observed between our two human labelers. In addition, we find a Pearson correlation of $r = 0.840$ ($p < 0.001$), suggesting a strong, positive linear relationship between the V-JEPA 2 classifier and the human labelers.

As second validation step, we use the classifier's labels to make direct comparisons to the real-world dVRK success rates. The results are presented in Figure~\ref{fig:cosmos-policy-eval-overview-b} and the per-task scores in Table~\ref{tab:policy_eval_quantity_all}. Most important for automated policy evaluation, we observe strong alignment with real-world policy ordering, with low disorder (Avg. MMRV$=0.10 \pm 0.04$). Moreover, a Pearson correlation of $r = 0.756$ ($p < 0.001$) indicates a strong, positive linear relationship between the Cosmos-Surg-dVRK with automated classifier success rates and the real-world dVRK Si success rates. Although the MBE rises to 0.153 versus 0.140 in human evaluations (see Table~\ref{tab:mbe}), further underscoring the inherent success bias observed earlier in the Cosmos-Surg-dVRK model, the overall statistical results closely match those of the human labeling experiment.

\begin{figure}[h]                
  \centering
  \includegraphics[width=0.85\linewidth]{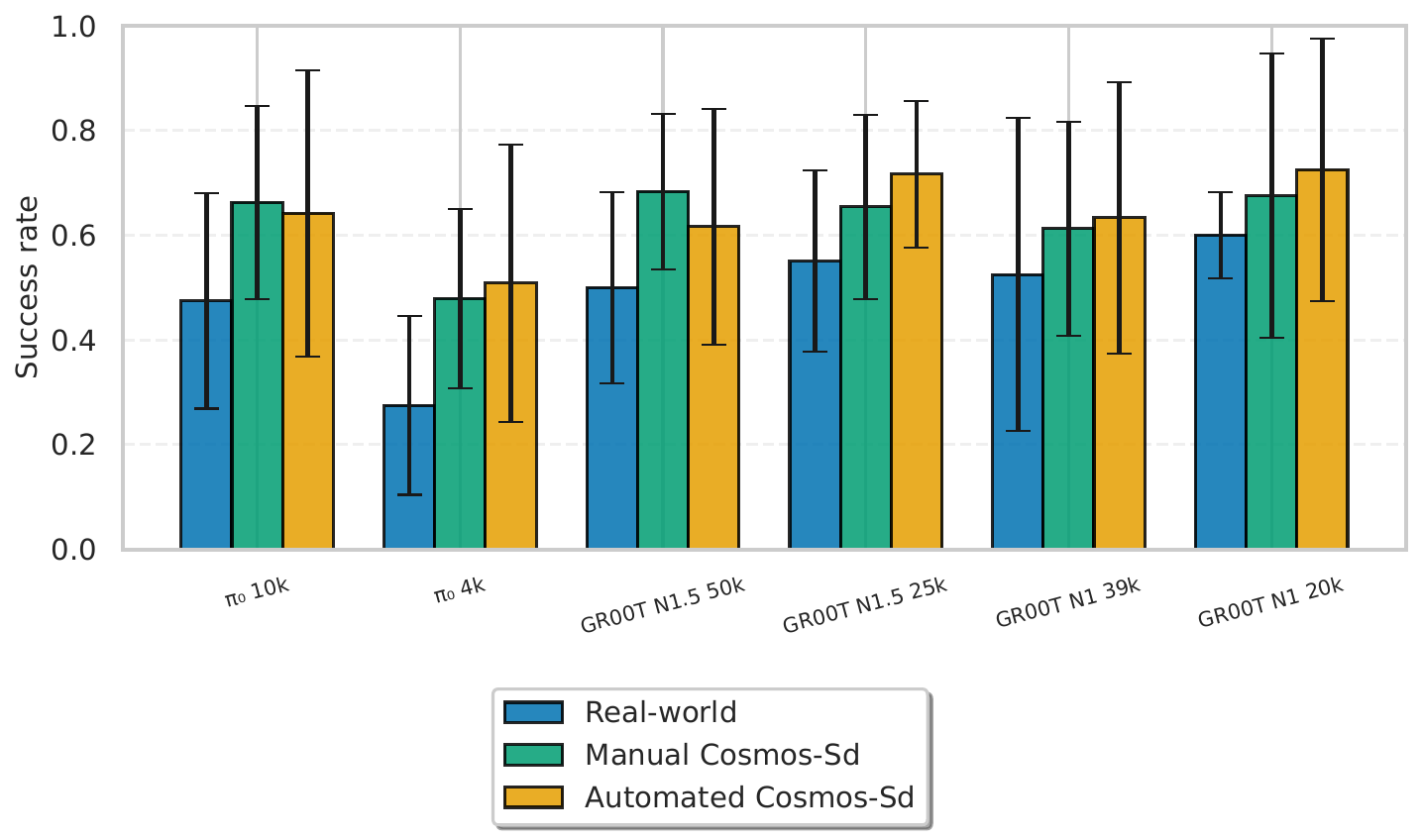}
  \caption{
    \textbf{Average tabletop policy success rates across evaluation techniques.} Results are averaged over four tasks. Each task includes 10 real-world rollouts and 10 Cosmos-Surg-dVRK rollouts (averaged over three seeds).
  }
  \label{fig:tt-policy-success-rates}
\end{figure}

The discussed inherent success bias of Cosmos-Surg-dVRK in both human evaluations and automated evaluations is visualized in Figure~\ref{fig:tt-policy-success-rates}. 

\subsubsection{On Cosmos-Surg-dVRK hallucinations}
While Cosmos-Surg-dVRK predictions correlate strongly with real dVRK Si rollouts, we observe discrepancies, which we term \textit{hallucinations}, in which Cosmos-Surg-dVRK depicts physically inconsistent outcomes. We attribute these hallucinations to insufficiently learned physics and kinematics due to limited distributional coverage in the training data, resulting in two primary error types:
\begin{itemize}
    \item False Positives: The model overestimates success. For example, it might show a needle incorrectly ``snapping'' into a misaligned gripper.
    \item False Negatives: The model underestimates success, such as showing a needle falling out of a gripper even when a perfect grasp is executed.
\end{itemize}
Qualitatively and quantitatively, we observe that Cosmos-Surg-dVRK has a bias towards false positives, as seen in Table~\ref{tab:mbe}. This tendency toward false positives reveals that, although Cosmos-Surg-dVRK correlates well with real outcomes, it has not yet sufficiently captured its understanding of the nuanced physical constraints involved in dVRK tabletop suturing tasks.

\subsubsection{On the importance of failure episodes}
Initially, we finetune Cosmos-Surg-dVRK on datasets containing only successful trajectories. Qualitatively, these models show a strong positive bias, often hallucinating false positives. We therefore hypothesize that including negative examples could mitigate this bias. To test this, we collect additional trajectories resulting in task failure (Figure~\ref{fig:tt_ds_composition}) and subsequently train two model variants: an ablated model on successes only, and a full model on both successes and failures.

Evaluations between these two finetunes are performed on holdout data unseen by both models. For each task and policy, two raters independently label 10 unique rollouts generated from a single seed, with the final success rate being the average of their assessments. The results are presented in Table~\ref{tab:policy_eval_quantity_all}. Across all tasks and metrics, the ablated Cosmos-Surg-dVRK without failure episodes yields lower performance, though the average MMRV of 0.146 is still comparable with the average MMRV of 0.129 of the full Cosmos-Surg finetune. However, as shown in Table~\ref{tab:mbe}, the MBE of 0.325 observed for manual success rate evaluation is significantly higher than that of the full Cosmos-Surg-dVRK version (MBE 0.140), suggesting that failure episodes represent a critical factor in surgical datasets for effective world model finetuning in surgical simulation.

\subsection{Ex-vivo Porcine Cholecystectomy Surgical Policy Evaluation in Cosmos-Surg-dVRK versus real dVRK}
Using the monocular version of the SRT-H policy~\citep{Kim2025SRTH}, we compare its performance in Cosmos-Surg-dVRK simulation with the real dVRK. The holdout test set has nine different porcine tissues and three tasks are performed, ``apply first clip'', ``apply third clip'', and ``cut the cystic duct''. Each task has three trials,  and as a consequence, we report majority-vote outcomes to ensure robustness to outlier judgments. This contrasts with the tabletop experiments, where the ten real-world trials enable us to report the average success rate, as the tabletop experiments' larger sample size allows stable statistical estimation. Thus, for the cholecystectomy experiments, each trial is rolled out three times using different seeds for Cosmos-Surg-dVRK. The generated videos are manually rated by two raters. Each trial is counted as a success if more than 50\% of the six labels are successful. A physics anomaly is counted as a failure only if it occurs as a proxy for poor policy action. The specific task success criteria are described in Appendix~\ref{app:chole_criteria}.

\begin{figure}[ht]                
  \centering
  \includegraphics[width=0.99\linewidth]{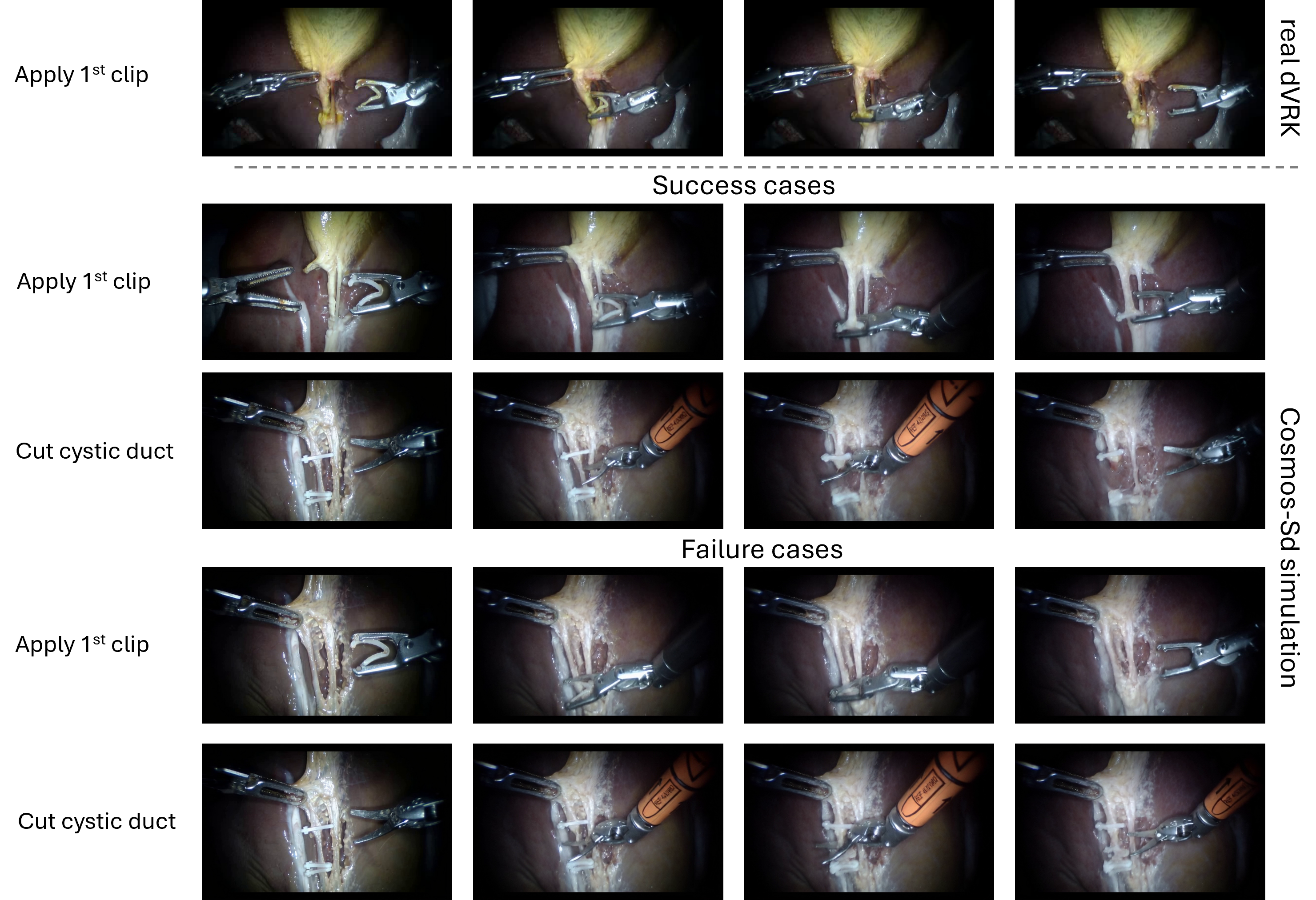}
  \caption{
    \textbf{Qualitative examples of SRT-H policy online evaluation in Cosmos-Surg-dVRK.} Each row represents one different policy rollout. Top row: Successful example as recorded on the real dVRK Si. Middle two rows: Successfully completed tasks in Cosmos-Surg-dVRK. Bottom two rows: Examples of failed tasks in Cosmos-Surg-dVRK. The four frames shown are chosen individually to best represent the example.
  }
  \label{fig:srth-policy-eval-qual-results}
\end{figure}

The exploratory results are summarized in Table~\ref{tab:srth_detailed}. We obtain an almost identical success rate overlap across nine trials. The causes of failure in the third rollout of the task ``apply third clip'' using Cosmos-Surg-dVRK include the scissor missing the duct or grabbing and cutting the wrong tube. Qualitative examples of success and failure cases are shown in Figure~\ref{fig:srth-policy-eval-qual-results}. 

\begin{table}[h]
\centering
\caption{SRT-H policy success rates on three cholecystectomy tasks.}
\label{tab:srth_detailed}
\begin{tabular}{lcccc}
\hline
\textbf{System} & \textbf{Apply first clip} & \textbf{Apply third clip} & \textbf{Cut cystic duct} & \textbf{Total} \\
\hline
dVRK & 3/3 & 2/3 & 2/3 & 7/9 \\
Cosmos-Surg-dVRK & 3/3 & 3/3 & 2/3 & 8/9\\
\hline
\end{tabular}
\end{table}

\subsubsection{Out-of-distribution trajectories on Cosmos-Surg-dVRK Porcine Cholecystectomy}
\begin{figure}[ht]                
  \centering
  \includegraphics[width=0.99\linewidth]{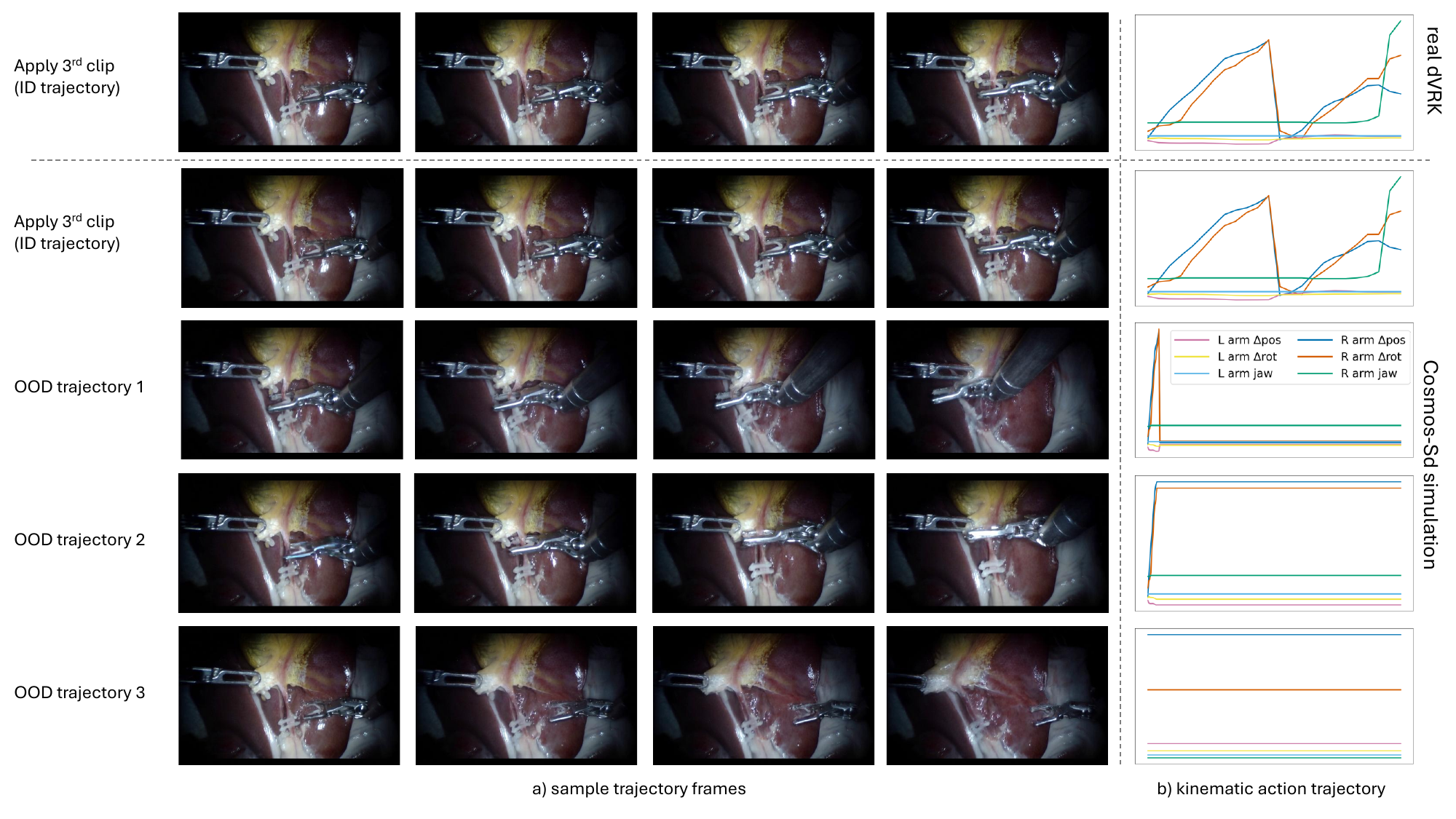}
  \caption{
    \textbf{Qualitative examples of cholecystectomy out-of-distribution action trajectories in Cosmos-Surg-dVRK.} Top row: four frames of an in-distribution (ID) trajectory rolled out on the real dVRK Si~\citep{Kim2025SRTH} with the corresponding trajectory plot. Second row: four frames of the identical trajectory rolled out in Cosmos-Surg-dVRK. Bottom three rows: Examples of out-of-distribution (OOD) trajectories not part of the training data distribution, rolled out in Cosmos-Surg-dVRK. a) shows four video frames chosen individually to best represent the example; b) visualization of the corresponding kinematic action trajectories of the two dVRK Si arms.
  }
  \label{fig:Cosmos-Surg-dVRK-ood-eval-qual-results}
\end{figure}

We experiment with Cosmos-Surg-dVRK on porcine cholecystectomy for use as a surgical simulator or surgical predictive assistant. To that end, we run Cosmos-Surg-dVRK inference with hand-crafted action trajectories to explore the generalization capabilities of the model to unseen action trajectory scenarios that are not part of the training data distribution. Qualitative results are presented in Figure~\ref{fig:Cosmos-Surg-dVRK-ood-eval-qual-results}.

\section{Limitations}
We finetune and evaluate the Cosmos WFM using a single camera and monocular policies. While state-of-the-art surgical policies use multi-view camera input~\citep{Kim2024SRT,Kim2025SRTH}, Cosmos-Surg-dVRK should be extended to support multi-view imaging. Most visuomotor policy architectures benefit from incorporating proprioceptive state information alongside images. In this study, Cosmos-Surg-dVRK exclusively models the visual state of each task, and all policies are trained without proprioceptive data. The generated videos often exhibit uncertainty at the boundaries of small objects such as needle, thread, and grippers, sometimes making success or failure judgments ambiguous. This ambiguity is plausibly partly aleatoric (e.g., occlusions, limited resolution) and partly epistemic, resulting from limited training data and the balance of success and failure episodes. While our initial ablation highlights the importance of failure episodes as a data category, further investigation is required to determine the optimal data composition for peak model performance.


\section{Conclusion}
In this work, we have introduced Cosmos-Surg-dVRK, an action-conditioned Cosmos WFM finetune for the rigorous online evaluation of surgical robot policy learning. Through a set of experiments on tabletop and ex-vivo cholecystectomy tasks, Cosmos-Surg-dVRK demonstrates a strong alignment with real dVRK robot performance, indicating its potential as a useful proxy for real-world benchmarking. Additionally, we propose the use of a V-JEPA 2 classifier that achieves strong correlation to human labelers and enables a fully automated policy evaluation pipeline for tabletop tasks. Our results suggest that Cosmos-Surg-dVRK facilitates more efficient and reproducible evaluation pipelines at reduced cost, while lowering some of the barriers associated with physical robot experimentation, supporting faster development and iteration of surgical autonomy approaches. Beyond policy benchmarking, Cosmos-Surg-dVRK enables ablation studies, which indicate that failure data are important for effective learning, and shows promise in generalizing to unseen action trajectories. Cosmos-Surg-dVRK’s ability to reproduce key aspects of real-world surgical scenarios demonstrates its potential value for both research and industry innovation, offering a learned simulator for exploring, refining, and evaluating new policy models prior to real-world deployment.

\bibliography{paper}
\bibliographystyle{paper}

\newpage
\appendix
\section{Appendix}

\subsection{V-JEPA 2 Attentive Probe Classifier: Implementation Details}
\label{app:vjepa_implementation_details}

\begin{figure}[htbp]
    \centering
    \includegraphics[width=0.95\textwidth]{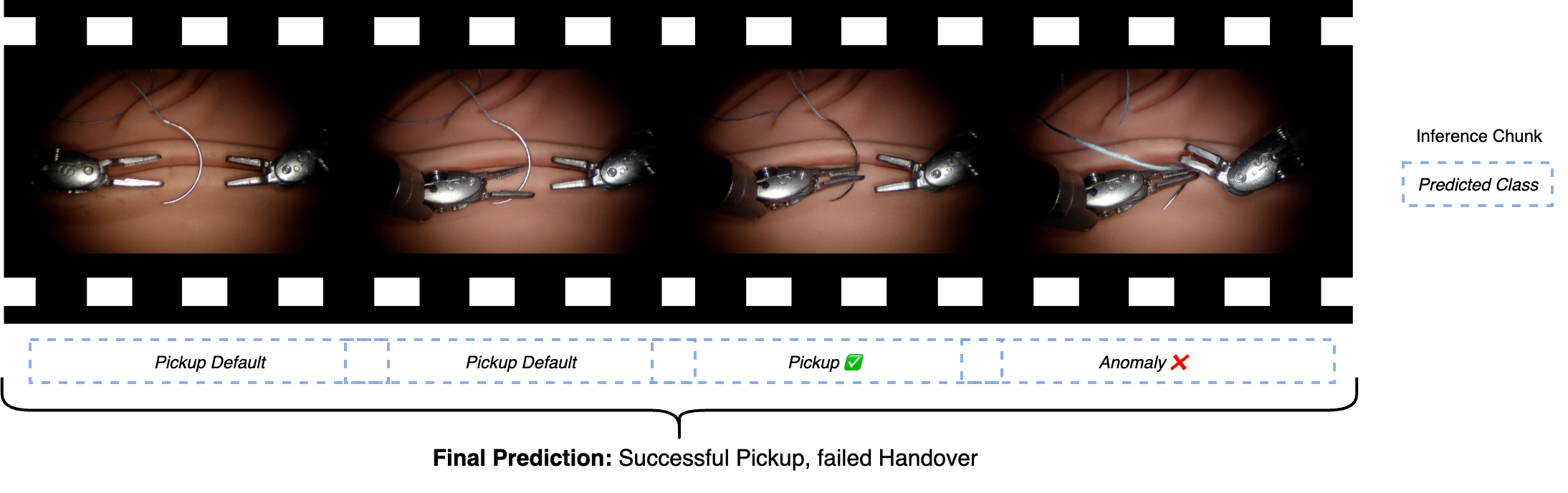}
    \caption{V-JEPA 2 Chunked Inference Methodology}
    \label{fig:vjepa-chunked-inference}
\end{figure}

To automate the labeling of failure and success events in our Cosmos-Surg-dVRK-produced policy rollouts, we train a 4-layer attentive probe classifier on top of frozen V-JEPA 2 video features. Due to the generative nature of our Cosmos WFM simulation environment, our policy rollout videos contain physics anomalies in addition to standard successes and failures. Consequently, for each tabletop task, our classifier utilizes three labels:
\begin{itemize}
    \item Success: The observed frames show the task being completed.
    \item Anomaly: The clip exhibits a physically irregular event, such as a gripper passing through an object.
    \item Default: The rollout is neither a success nor an anomaly.
\end{itemize}

 To fit within the context length of V-JEPA 2, we chunk the initial videos that are of hundreds of frames into 32 frame segments with 6 frames of overlap and sequentially classify each video chunk to produce an overall label for each policy rollout. The final rollout classification is derived as follows:

\begin{itemize}
    \item Success occurs before anomaly event $\rightarrow$ \textbf{success}
    \item Anomaly event occurs before success $\rightarrow$ \textbf{failure}
    \item Success event not detected $\rightarrow$ \textbf{failure}
\end{itemize}
 
To train this classifier, we manually labeled 2,310 32-frame clips for the categories described above. Due to the importance of the physics anomaly events, we augment standard cross-entropy loss with loss scaling factors for false-positive and false-negative anomaly events during training. We perform a hyperparameter search using the BOHB algorithm~\cite{falkner2018bohbrobustefficienthyperparameter}, with the average validation F1-score as our optimization target. The search includes the loss scaling factors, learning rate, batch size, and the number of warm-up steps. Complete training details are included in Table~\ref{tab:vjepa_training_details}.

\begin{table}[h]
\centering
\caption{V-JEPA Training Configuration}
\label{tab:vjepa_training_details}
\begin{tabular}{lr}
\toprule
\textbf{Parameter} & \textbf{Value} \\
\midrule
\multicolumn{2}{l}{\textit{Backbone (Frozen)}} \\
Model Architecture & ViT-Huge \\
Backbone Parameters & $\sim$632M \\
Embedding Dimension & 1280 \\
Depth & 32 \\
Attention Heads & 16 \\
\midrule
\multicolumn{2}{l}{\textit{Classifier (Trainable)}} \\
Architecture & Attentive Pooler \\
Classifier Depth & 4 blocks \\
Classifier Heads & 16 \\
Classifier Parameters & $\sim$77M \\
Number of Classes & 11 \\
\midrule
\multicolumn{2}{l}{\textit{Training Configuration}} \\
Learning Rate & 2.85e-5 \\
Training Hz & 10 Hz \\
Batch Size & 8 \\
Number of Epochs & 200 \\
Warmup Epochs & 5 \\
False Positive Penalty & 5.0 \\
Anomaly Penalty & 1.0 \\
\bottomrule
\end{tabular}
\end{table}

\begin{table}[h]
\centering
\caption{Cosmos-Surg-dVRK Training Configuration.}
\label{tab:training_config}
\begin{tabular}{ll}
\toprule
\textbf{Parameter} & \textbf{Value} \\
\midrule
\multicolumn{2}{l}{\textit{Model Configuration}} \\
Architecture & Cosmos-Predict2-2B \\
Pretrained Checkpoint & Cosmos-Predict2-2B-Video2World \\
Parallelism Strategy & FSDP \\
Context Parallel Size & 1 \\
\midrule
\multicolumn{2}{l}{\textit{Training Configuration}} \\
Number of Nodes & 4 \\
GPUs per Node & 8 \\
Total GPUs & 32 \\
Batch Size per GPU & 24 \\
Effective Global Batch Size & 768 \\
\midrule
\multicolumn{2}{l}{\textit{Optimization}} \\
Optimizer & FusedAdamW \\
Learning Rate & $2.4 \times 10^{-4}$ \\
Scheduler & LambdaLinear \\
Warmup Steps & 1,000 \\
Total Training Steps & 20,000 \\
$f_{\text{max}}$ & 1.0 \\
$f_{\text{min}}$ & 0.1 \\
\bottomrule
\end{tabular}
\end{table}

\subsection{Cholecystectomy Task Manual labeling Success criteria}
\label{app:chole_criteria}
The following criteria define a successful outcome for each of the autonomous cholecystectomy surgical tasks. A trial is marked as a success only if all success criteria for that task are met. Should a physics anomaly be observed that directly impacts the outcome of the attempted task, that rollout is marked as a failure.
\begin{itemize}
  \item Task 'apply 1st clip': The left arm pulls the gallbladder. The right arm is moved to the left tube (cystic duct) and the clip applier embraces the left tube. The applier is closed and the clip mounted on the tube. The applier opens again and the clip stays on the tube. The right arm pulls back to its original position and finally the left arm slightly lowers the tension on the gallbladder.
  \item Task 'apply 3rd clip': The left arm pulls the gallbladder. The right arm is moved to the left tube (cystic duct), slightly above the two applied clips, the clip applier embraces the left tube. The right arm moves up along the tube to create a space between second and third clip. The applier is closed and the clip mounted on the tube. The applier opens again and the clip stays on the tube. The right arm pulls back to its original position and finally the left arm slightly lowers the tension on the gallbladder.
  \item Task 'cut cystic duct': The left arm pulls gallbladder, right arm moves scissor in between third clip and second clip, right arm pulls the cystic duct away a few millimeters, and finally the scissors close to cut the duct.
\end{itemize}

\subsection{Tabletop Task Manual labeling Success criteria}
The following criteria define a successful outcome for each of the four tabletop surgical tasks. A trial is marked as a success only if all success criteria for that task are met. Should a physics anomaly be observed that directly impacts the outcome of the attempted task, that rollout is marked as a failure.

\begin{itemize}
  \item Task 'needle pickup': The left arm approaches the needle on the suture pad. Next, one gripper jaw presses into the pad while the arm simultaneously moves to the right, scooping the needle between the jaws. Finally, the gripper closes to secure the needle.
  \item Task 'needle handover': Following a successful needle pickup, the left arm raises the secured needle above the suture pad. The right arm then maneuvers to position its gripper around the raised needle. Once the right gripper closes to secure the needle, the left gripper releases it. The right arm then retracts with the secured needle.
  \item Task 'needle throw': The right arm begins with the needle firmly secured in its gripper. The right arm then navigates the needle tip to the back wall of the suture pad. Once well positioned, the arm drives the needle downwards and twists the needle towards the front wall of the suture pad, piercing the back wall and pushing the needle through to the front wall.
  \item Task 'knot tying': The right arm begins with the needle grasped and the suture passed through the tissue. First, the right arm lifts the needle, creating a span of thread that the left arm moves behind. The right arm then performs a circular motion to wrap the suture around the left arm's gripper, forming a loop. Next, the left arm reaches through this loop to grasp the free tail of the suture. With the tail secured, both arms pull outwards in opposite directions. This movement pulls the tail through the loop, which slides off the gripper and tightens down, forming a secure knot at the center of the suture pad.
\end{itemize}

\newpage
\subsection{Bland-Altman analysis of Cosmos-Surg-dVRK without failure episodes}
\begin{figure}[h]
    \centering
    \includegraphics[width=0.6\linewidth]{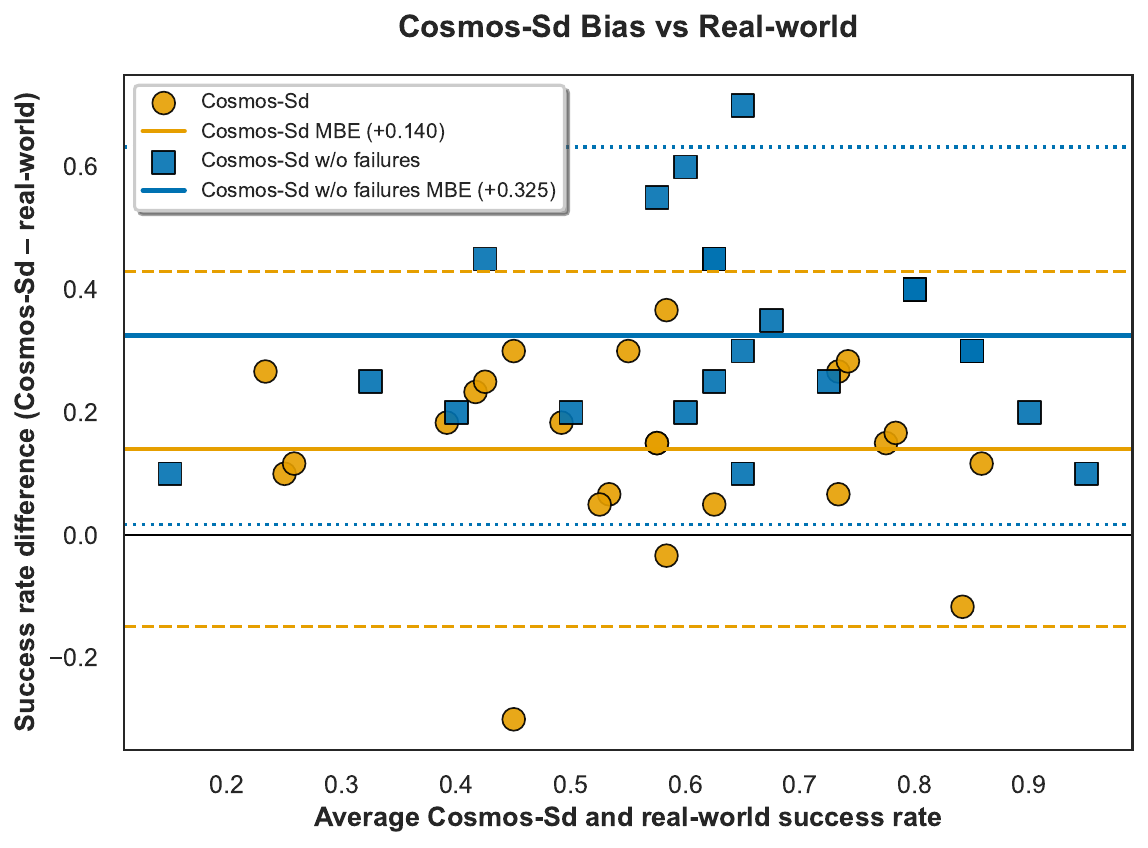}
    \caption{
        \textbf{Bland–Altman analysis.}
        Mean bias error (MBE) and limits of agreement in Cosmos-Surg-dVRK vs.\ Cosmos-Surg-dVRK without failure episodes across all policies and tasks.
    }
    \label{fig:ba_success_bias}
\end{figure}

Figure~\ref{fig:ba_success_bias} visualizes the positive shift in higher success rates between the Cosmos-Surg-dVRK without failure episodes version and the full Cosmos-Surg-dVRK full version. The ablated Cosmos-Surg-dVRK model exhibits a markedly larger positive mean bias error, and limits of agreement that are shifted upward compared to the full Cosmos-Surg-dVRK model.

\newpage
\subsection{Additional qualitative examples of tabletop policy online rollouts in Cosmos-Surg-dVRK}
\label{app:tt_add_quali_examples}
\begin{figure}[ht]                
  \centering
  \includegraphics[width=0.99\linewidth]{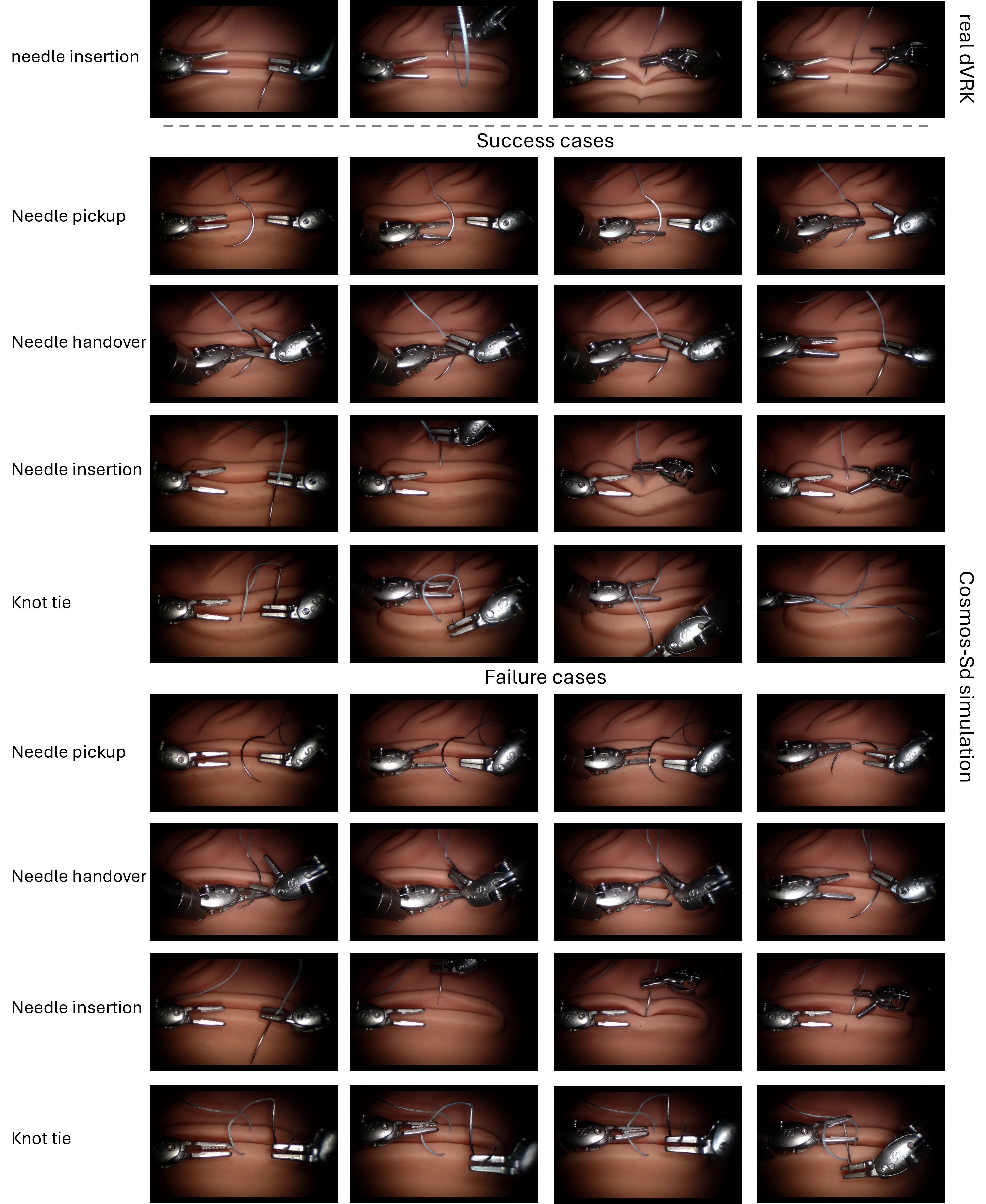}
  \caption{
    \textbf{Qualitative examples of tabletop policy online rollouts in Cosmos-Surg-dVRK.} Each row represents a different policy rollout. Top row: Successful example as recorded on the real dVRK. Rows 2 - 5: Successfully completed tasks in Cosmos-Surg-dVRK. Bottom four rows: Examples of failed tasks in Cosmos-Surg-dVRK. The four frames shown are chosen individually to best represent the example.
  }
  \label{fig:tt_add_quali_examples_app}
\end{figure}

\newpage
\subsection{Additional qualitative examples of ex-vivo porcine cholecystectomy policy online rollouts in Cosmos-Surg-dVRK}
\label{app:chole_add_quali_examples}
\begin{figure}[ht]                
  \centering
  \includegraphics[width=0.99\linewidth]{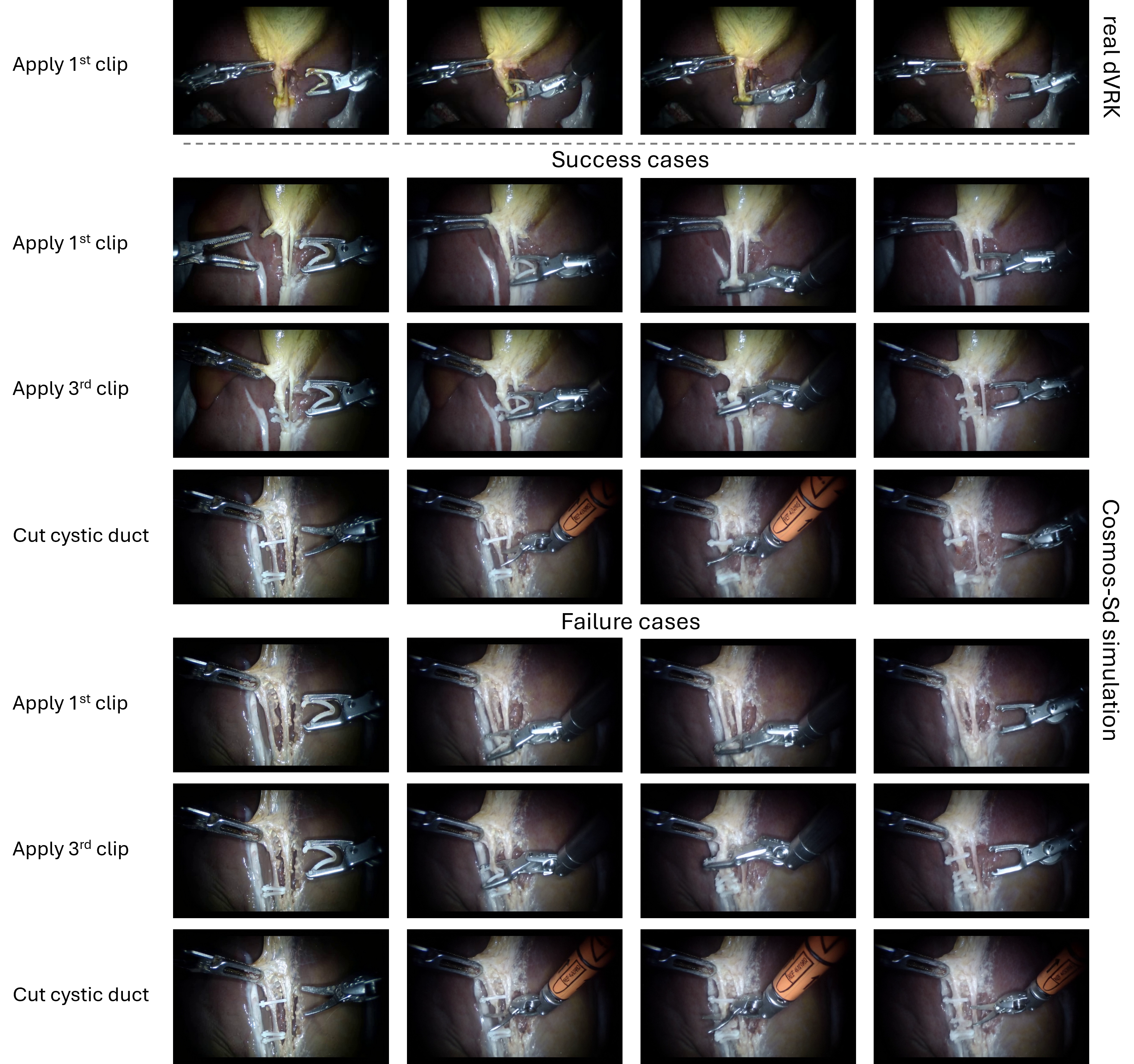}
  \caption{
    \textbf{Qualitative examples of SRT-H policy online evaluation in Cosmos-Surg-dVRK.} Each row represents one different policy rollout. Top row: Successful example as recorded on the real dVRK. Middle three rows: Successfully completed tasks in Cosmos-Surg-dVRK. Bottom three rows: Examples of failed tasks in Cosmos-Surg-dVRK. The four frames shown are chosen individually to best represent the example.
  }
  \label{fig:chole_add_quali_examples_app}
\end{figure}
\end{document}